\documentclass{article}

\PassOptionsToPackage{numbers, compress}{natbib}
\usepackage[preprint]{neurips_2026}

\usepackage[utf8]{inputenc} 
\usepackage[T1]{fontenc}    
\usepackage{hyperref}       
\usepackage{url}            
\usepackage{booktabs}       
\usepackage{amsfonts}       
\usepackage{nicefrac}       
\usepackage{microtype}      
\usepackage[table]{xcolor}         
\usepackage{multirow}
\usepackage{comment}
\usepackage{amsmath}    

\usepackage{graphicx}

\title{GenRecon: Bridging Generative Priors for Multi-View 3D Scene Reconstruction}

\author{%
  Katharina Schmid\textsuperscript{1} \quad
  Nicolas von Lützow\textsuperscript{1} \quad
  Jozef Hladký\textsuperscript{2} \quad
  Angela Dai\textsuperscript{1} \quad
  Matthias Nie{\ss}ner\textsuperscript{1} \\[0.6em]
  \normalsize
  \textsuperscript{1}Technical University of Munich \quad
  \textsuperscript{2}Computing Systems Lab, Huawei Technologies, Switzerland
}

\begin{document}

\maketitle

\vspace{-0.5cm}

\begin{figure}[h]
    \centering
    \includegraphics[width=1.0\linewidth]{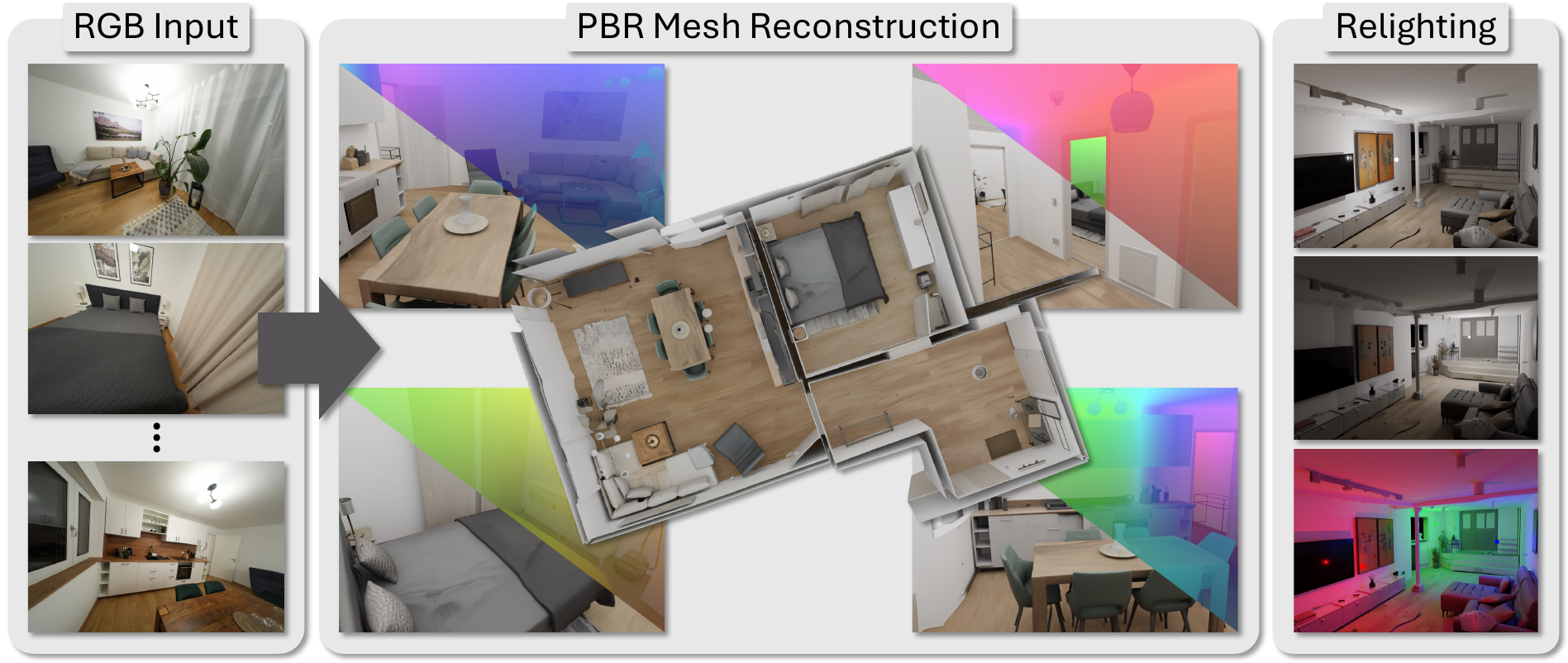}
    \caption{\textbf{GenRecon.} Given a sparse set of RGB images of an indoor scene (left), our method reconstructs a complete, high-fidelity PBR mesh (center) by formulating scene reconstruction as conditional 3D generation over overlapping spatial chunks. The recovered mesh with material properties enables realistic relighting and editing in standard rendering pipelines (right).}
    \label{fig:my_image}
\end{figure}

\begin{abstract}
We introduce a new approach to high-fidelity 3D scene reconstruction from multi-view RGB images that tightly couples reconstruction with a strong generative 3D prior. 
We cast scene reconstruction as conditional 3D generation over a set of spatially-localized, overlapping chunks that together tile the scene, scaling generation to large scene extents.
Crucially, we inherit the fidelity and completeness of state-of-the-art generative shape models -- we use Trellis.2 as an example -- which we generalize to the scene level.
To this end, we propose a projection-based conditioning mechanism that lifts posed multi-view image features into a coherent 3D representation aligned with the generative model, independent of view ordering and spatially anchored to the scene, yielding high-fidelity, multi-view consistent generated geometry. 
This enables lifting the strong object-level prior of Trellis.2 to multi-view, scene-scale generation, producing faithful, editable PBR mesh reconstructions of indoor environments. 
As a result, we obtain high-fidelity results that outperform cutting-edge reconstruction methods by 16\%.
\end{abstract}

\section{Introduction}

Reconstructing high-quality 3D scenes from multi-view RGB images is a fundamental problem in computer vision and graphics, underpinning applications ranging from AR/VR and robotics to embodied AI, simulation, and digital content creation. 
For instance, a robot navigating a cluttered environment, an artist importing a captured environment into a game engine, and an immersive VR experience transporting a user to a distant real-world setting can all be powered by 3D scene reconstructions. 
The requirements imposed on a reconstruction, however, vary  across these settings.
For navigation and perception, reconstruction primarily provides the geometric structure needed for downstream tasks, where metric accuracy is prioritized and high surface and visual fidelity is not essential.
For content creation and immersive applications, 3D reconstructed scenes must meet a substantially higher fidelity bar, matching the quality of crafted 3D assets with complete, high-fidelity surfaces along with material properties suitable for relighting and editing. 

Achieving such high-fidelity 3D scene reconstruction from multi-view images is fundamentally challenging, as it is a highly underconstrained inverse task. 
From only a set of 2D views, recovering the actual 3D structure at any given location requires many observations from diverse viewpoints. This requires reliable correspondences to be established, a difficult problem due to needing both sophisticated appearance and semantic understanding to handle textureless regions, repetitive patterns, large viewpoint changes, and view-dependent effects. Real scene captures rarely satisfy diverse, accurate correspondences everywhere in a scene, so per-scene optimization-based approaches \cite{li2023neuralangelohighfidelityneuralsurface,wang2023neuslearningneuralimplicit,yariv2021volumerenderingneuralimplicit,yu2022monosdfexploringmonoculargeometric,chen2025neusgneuralimplicitsurface,kerbl20233dgaussiansplattingrealtime,huang20252dgaussiansplattinggeometrically,chen2025pgsrplanarbasedgaussiansplatting} often produce incomplete, noisy, or oversmoothed reconstructions in these underconstrained regions.

Despite these challenges, recent works have made significant progress in reconstruction by incorporating learned priors. Feed-forward scene reconstruction methods \cite{wang2024dust3rgeometric3dvision,leroy2024groundingimagematching3d,wang2025vggtvisualgeometrygrounded,lin2025depth3recoveringvisual,sun2021neuralreconrealtimecoherent3d,stier2023finerecondepthawarefeedforwardnetwork,charatan2024pixelsplat3dgaussiansplats,chen2024mvsplatefficient3dgaussian,jiang2025anysplatfeedforward3dgaussian} have transformed the field, recovering geometry directly from images in a single pass and producing remarkably consistent reconstructions that have the potential to power downstream navigation and perception tasks. Unfortunately, their outputs remain ill-suited to the required fidelity needed for content creation scenarios: surfaces remain noisy or oversmoothed in challenging regions, and incomplete in occluded and unobserved areas. 
At the same time, generative 3D modeling has made rapid strides in producing realistic, coherent, and complete 3D object shapes. Modern generative shape models \cite{xu2024instantmeshefficient3dmesh,liu2023one2345singleimage3d,liu2024syncdreamergeneratingmultiviewconsistentimages,zhao2025hunyuan3d20scalingdiffusion,liao2025completegaussiansplatssingle,chang2025reconviagenaccuratemultiview3d} capture powerful priors over high-quality shape geometry, enabling the synthesis of detailed, structurally consistent 3D assets. 
We observe that these strong generative 3D shape priors offer a powerful opportunity in scene reconstruction: by integrating strong generative shape priors directly into multi-view reconstruction, this enables reconstruction of complete, high-fidelity 3D scene assets. 

In this work, we introduce a new approach that tightly couples multi-view 3D reconstruction with a strong generative 3D prior. We formulate scene reconstruction as conditional 3D generation over a set of spatially-localized, overlapping scene chunks, enabling large-scale reconstruction while inheriting the fidelity, completeness, and realism of state-of-the-art generative shape model Trellis.2 \cite{xiang2025nativecompactstructuredlatents}. We recast the shape generative prior from Trellis.2 to support multi-view scene chunk generation by formulating a projection-based conditioning pathway that injects posed multi-view image information into the generative model in a spatially-grounded, permutation-invariant manner. This allows precise control over both generated geometry and spatial alignment. By preserving the pretrained prior through parameter-efficient fine-tuning on synthetic scene data, our method produces faithful, editable PBR mesh reconstructions of indoor scenes, significantly narrowing the gap between current reconstruction capabilities and the quality required for content creation scenarios. 
To summarize, our contributions are:
\begin{itemize}
    \item We introduce a new approach for reconstructing scene-level PBR meshes from RGB images, by coupling multi-view reconstruction with a strong object-level 3D generative prior. We formulate reconstruction as conditional 3D generation over overlapping spatial scene chunks, casting scene recovery as a single coherent generation process in which all chunks are jointly synthesized under the guidance of the input views.
    \item To enable this, we extend a single-image, object-level 3D generative prior to a multi-image, pose-controlled setting via a dedicated 3D conditioning pathway. This pathway lifts multi-view image features using explicit camera poses and fuses them in a spatially-grounded, permutation-invariant manner, enforcing strict geometric consistency across views while enabling precise control over the resulting 3D structure.
\end{itemize}

\section{Related Work}

\paragraph{Reconstruction without learned priors.}
Classic multi-view stereo pipelines such as COLMAP~\cite{schonbergerStructurefromMotionRevisited2016} reconstruct geometry through feature matching, epipolar verification, and patch-based stereo fusion, but relying solely on photoconsistency, they cannot recover structure in weakly textured, occluded, or sparsely observed regions. Neural implicit surface methods  \cite{li2023neuralangelohighfidelityneuralsurface,wang2023neuslearningneuralimplicit,yariv2021volumerenderingneuralimplicit,mildenhall2020nerfrepresentingscenesneural} extend this paradigm by representing scenes as continuous signed-distance or density fields optimized via differentiable volume rendering; while they recover smoother surfaces, they still fail in ambiguous regions where triangulation is under-constrained. MonoSDF~\cite{yu2022monosdfexploringmonoculargeometric} and NeuSG~\cite{chen2025neusgneuralimplicitsurface} attempt to mitigate these limitations by augmenting per-scene optimization with monocular depth cues or jointly optimized 3D Gaussian Splatting guidance, yet both remain unable to generate geometry beyond what photoconsistency can constrain. Similarly, Gaussian splatting methods \cite{kerbl20233dgaussiansplattingrealtime,huang20252dgaussiansplattinggeometrically,chen2025pgsrplanarbasedgaussiansplatting} optimize explicit anisotropic primitives via differentiable rasterization for fast rendering, but without learned shape priors, they suffer from the same noise and incompleteness in unobserved regions. MeshSplats~\cite{tobiasz2026meshsplatsmeshbasedrenderinggaussian} translates these primitives into disjoint mesh faces for standard ray-tracing pipelines, improving editability but inheriting the underlying reconstruction artifacts.

\paragraph{Reconstruction with learned priors.}
The shift toward learned priors has produced geometric foundation models~\cite{wang2024dust3rgeometric3dvision,leroy2024groundingimagematching3d,wang2025vggtvisualgeometrygrounded,wang2025continuous3dperceptionmodel,lin2025depth3recoveringvisual} that regress dense depth maps, pointmaps, or camera parameters directly from images. While these achieve impressive geometric recovery on observed surfaces, their unstructured outputs must be fused into surfaces post hoc, and they lack generative priors to complete occluded regions. Building on this direction, feed-forward volumetric fusion methods~\cite{sun2021neuralreconrealtimecoherent3d,stier2023finerecondepthawarefeedforwardnetwork,božič2021transformerfusionmonocularrgbscene,stier2021vortxvolumetric3dreconstruction,gao2023visfusionvisibilityawareonline3d,na2024uforecongeneralizablesparseviewsurface,sayed2022simplerecon3dreconstruction3d} backproject image features into 3D volumes to directly regress TSDF or occupancy fields, while feed-forward Gaussian splatting methods~\cite{charatan2024pixelsplat3dgaussiansplats,chen2024mvsplatefficient3dgaussian,jiang2025anysplatfeedforward3dgaussian,xu2025depthsplatconnectinggaussiansplatting,ye2024poseproblemsurprisinglysimple,hong2025pf3platposefreefeedforward3d,ye2025yonosplatneedmodelfeedforward,wang2025freesplatgeneralizable3dgaussian} localize explicit primitives in a single forward pass. Despite learning strong geometric priors, all of these approaches remain deterministic regressors that cannot generate coherent geometry in unobserved regions, and they produce unstructured depth maps or Gaussian clouds rather than editable meshes.
To move beyond deterministic regression, recent methods have explored generative priors. Approaches based on 2D and video diffusion \cite{gao2024cat3dcreate3dmultiview,wu2023reconfusion3dreconstructiondiffusion,guo2025multiviewreconstructionsfmguidedmonocular,hu2024depthcraftergeneratingconsistentlong,xu2025geometrycrafterconsistentgeometryestimation,liu2025reconxreconstructscenesparse,yu2024viewcraftertamingvideodiffusion} synthesize intermediate frames or depth maps that are then reconstructed into 3D via downstream fusion or per-scene optimization, rather than directly generating structured geometry. 
Native 3D generative models \cite{xu2024instantmeshefficient3dmesh,liu2023one2345singleimage3d,liu2024syncdreamergeneratingmultiviewconsistentimages,zhao2025hunyuan3d20scalingdiffusion} come closer to direct 3D output, but largely operate at the object level and are typically restricted to single-view conditioning.
MV-SAM3D~\cite{li2026mvsam3dadaptivemultiviewfusion} and ReconViaGen~\cite{chang2025reconviagenaccuratemultiview3d} move toward multi-view conditioning, but still operate at object level.
Concurrent to our work, Pixal3D~\cite{li2026pixal3dpixelaligned3dgeneration} adopts a closely related conditioning strategy, back-projecting multi-scale image features into a 3D feature volume to establish explicit pixel-to-3D correspondence and thereby natively supporting pose-controlled single- and multi-view inputs.
However, it remains restricted to object-level generation and does not produce PBR-textured geometry.
DiffusionGS~\cite{liao2025completegaussiansplatssingle} extends generative completion to scenes yet conditions on a single image and outputs unstructured Gaussian splats.
Compositional methods \cite{li2026mvsam3dadaptivemultiviewfusion,li2026pixal3dpixelaligned3dgeneration,zhang2023scenewiz3d,zhou2024gala3d,liDISceneObjectDecoupling2024,chen2024comboverse,han2025reparo,yao2025cast,dominici2025dreamanywhere} decompose inputs into individual objects, reconstruct each with an off-the-shelf generative model, and assemble them via post-hoc layout optimization.
While this paradigm leverages strong object priors, it decouples generation from composition: object boundaries may be inconsistent, occluded geometry is hallucinated independently, and inter-object relations are enforced by optimization rather than emerging from a single coherent generative process.

In contrast to these approaches, our method leverages a pretrained 3D generative prior to directly synthesize complete, structured mesh geometry conditioned on the input views. By formulating scene reconstruction as a single coherent conditional 3D generation process over overlapping spatial chunks, we bypass per-object decomposition, per-view fusion, and per-scene optimization entirely.

\section{Method}

\begin{figure}[h]
    \centering
    \includegraphics[width=1\linewidth]{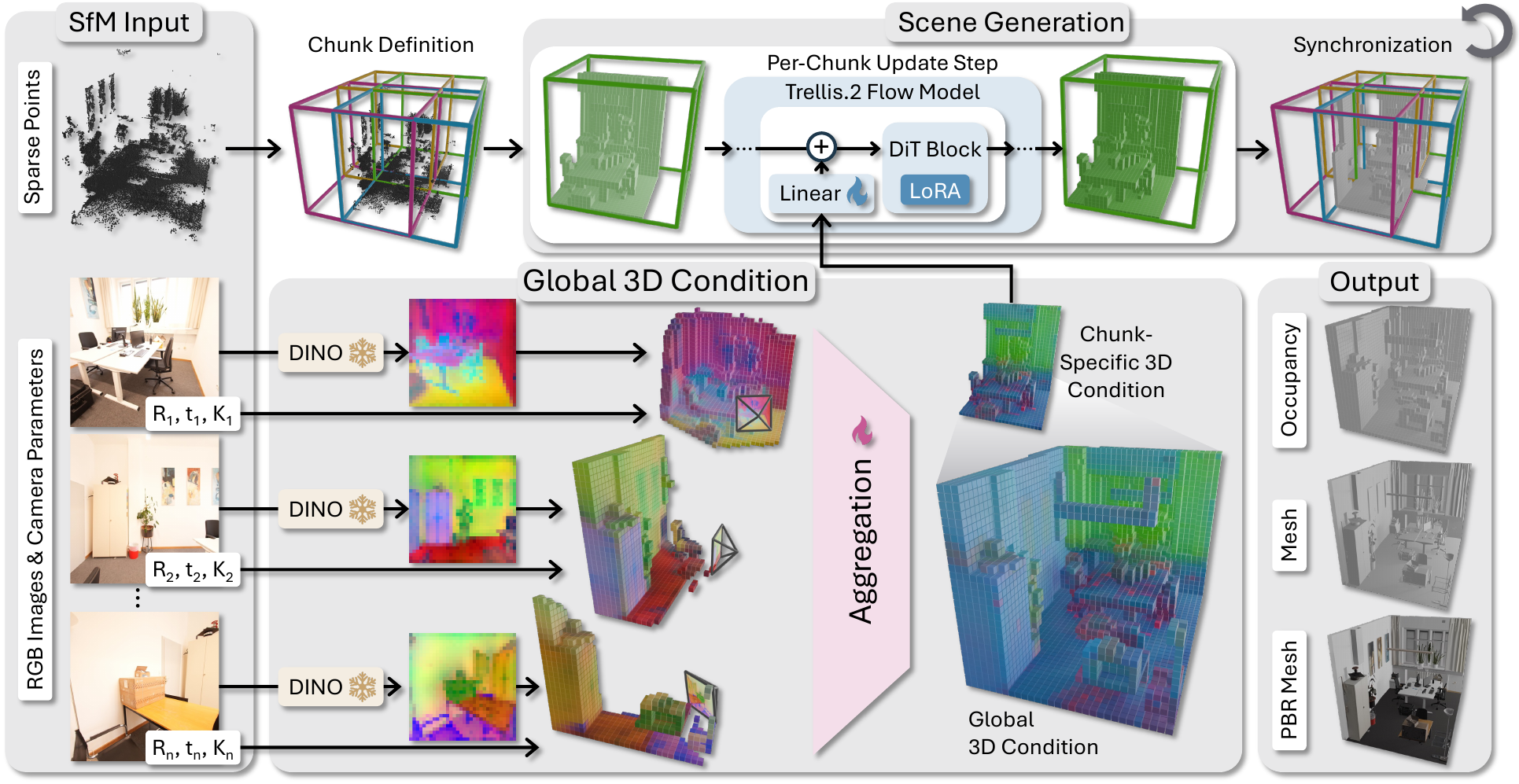}
    \caption{\textbf{Pipeline overview.} Given posed RGB images and a sparse point cloud from SfM (left), we define overlapping scene chunks and construct a global 3D conditioning grid by lifting DINOv3 image features into per-view volumes and aggregating them (center).
    By extending a 3D generative prior with a new spatially-grounded multi-view conditioning pathway, we jointly generate all chunks in a single flow-matching trajectory to recover a complete, pose-aligned scene-level PBR mesh (right).}
    \label{fig:pipeline}
\end{figure}

We address the problem of reconstructing a complete, high-fidelity 3D scene from a sparse, unordered set of \( N \) posed RGB images \( \{I_n\}_{n=1}^N \) with associated camera intrinsics and extrinsics \( \{K_n, T_n\}_{n=1}^N \). Our output is a scene-level mesh \( \mathcal{M} \) with PBR materials, suitable for direct integration into rendering and authoring pipelines (Figure~\ref{fig:pipeline}).
Our approach tightly couples multi-view reconstruction with a strong generative 3D prior by casting scene reconstruction as the joint generation of a set of overlapping scene chunks that cover the entire scene (Section \ref{sec:pipeline}).
To realize this, we employ the object-level generative prior from Trellis.2~\cite{xiang2025nativecompactstructuredlatents}, and recast it for  scene-level generation by introducing  multi-view conditioning that spatially grounds scene chunk generation  (Section \ref{sec:prior}).

\subsection{Multi-view scene chunk generation}
\label{sec:prior}

\paragraph{Scene chunks.}
We define a scene chunk \( c \) as a fixed-size 3D volume \( V_c = [0, L]^3 \) in its own canonical coordinate frame, paired with a translation \( t_c \in \mathbb{R}^3 \) that places it in the world frame. 
Each chunk \( c \) is associated with a set of input views \( \mathcal{V}_c \subseteq \{1, \dots, N\} \) whose cameras observe \( c \).
Our generative model $\Phi$ takes as input the chunk's canonical volume specification and the posed views \( \{(I_n, K_n, T_n^{-1} T_c)\}_{n \in \mathcal{V}_c} \), where \( T_c\) denotes the chunk-to-world transform corresponding to \( t_c \), and produces a 3D latent \( z^{(c)} \) representing the geometry and appearance within \( V_c \).

\paragraph{Generative prior.}
We instantiate our generative model \( \Phi \) from Trellis.2~\cite{xiang2025nativecompactstructuredlatents}, a state-of-the-art 3D shape generative model that produces high-quality objects by first predicting coarse occupancy, followed by high-fidelity shape and PBR texture.
These are parameterized by a flow-matching \cite{lipman2023flowmatchinggenerativemodeling} denoiser operating on the respective latent features. 
Trellis.2 is designed to take a single unposed RGB image as input through cross-attention; the position, orientation, and scale of generated content are not specified by the input but implicitly determined by the model's training distribution. While this design enables high-quality object generation, the single-image, pose-free conditioning regime is ill-suited for scene reconstruction: capturing large scenes inherently requires multiple views that the model must consume as a coherent set, as well as place generated content in a known coordinate frame so that adjacent chunks compose consistently.

\paragraph{Spatially-grounded multi-view conditioning.}
We address both gaps with a single design: a 3D conditioning pathway that carries multi-view image evidence into the generative model \( \Phi \) in a spatially anchored, view-order-invariant form. Given a chunk \( c \) with associated views \( \mathcal{V}_c \), we encode each image independently, lift the resulting per-view features into 3D grids over the chunk's volume, and aggregate across views in a permutation-invariant fashion to obtain the chunk's 3D conditioning \( G^{(c)} \).

We encode each input image with DINOv3~\cite{siméoni2025dinov3}, producing a dense 2D feature map \(F_n\) 
for each view to keep this input distribution close to Trellis.2's pretraining. 
For each view \( n \in \mathcal{V}_c \), we then lift \( F_n \) into a per-view 3D feature grid \( G_n^{(c)} \) defined over the chunk's canonical volume. Each voxel \( x \in V_c \) is projected into the view's image plane via \( \pi_n(x) = K_n T_n^{-1} (x + t_c) \), and the corresponding feature is retrieved: 
\(G_n^{(c)}(x) = F_n\bigl(\pi_n(x)\bigr) \).
This projection step spatially grounds the design, tying every conditioning feature to an explicit 3D location in the chunk's coordinate frame.

Finally, the per-view grids \( \{G_n^{(c)}\}_{n \in \mathcal{V}_c} \) are aggregated into a single 3D conditioning grid \( G^{(c)} \) using an IBRNet-style~\cite{wang2021ibrnet} scheme. The aggregation is permutation-invariant across views and for arbitrary~\( |\mathcal{V}_c| \), enabling our approach to handle variable numbers of input images without needing a canonical view ordering:
For each voxel with per-view features \( \{f_i\}_{i=1}^N \), we first compute
cross-view statistics \( \mu = \tfrac{1}{N}\sum_i f_i \) and \( \sigma^2 = \tfrac{1}{N}\sum_i f_i^2 - \mu^2 \), which serve as global context shared across views. Each view's feature is refined and assigned an aggregation logit by two small MLPs sharing the same input:
\begin{equation}
f'_i = \mathrm{MLP}_\text{feat}([f_i, \mu, \sigma^2]),
\qquad
w_i = \mathrm{MLP}_\text{weight}([f_i, \mu, \sigma^2]).
\end{equation}
The final voxel feature is the mean plus a softmax-weighted residual, where $\mathrm{MLP}_\text{feat}$'s final layer is zero-initialized so the module starts training as a cross-view mean:
\begin{equation}
f_\text{out} = \mu + \sum_i \alpha_i\, f'_i,
\qquad 
\text{where}
\enspace
\alpha_i = \mathrm{softmax}_i(w_i).
\end{equation}

\paragraph{Conditioning injection.}
The aggregated 3D condition \( G^{(c)} \) is injected into the generative denoiser~\( \Phi \) residually at each block, added voxel-wise through a zero-initialized layer so that initialization preserves the pretrained model's behavior. Because \( G^{(c)} \) is defined directly on the chunk's coordinate frame, every conditioning signal carries explicit positional meaning, and view consistency and pose control fall out as direct consequences of the design rather than properties the model must learn.

\paragraph{Training.}
We train the conditioning pathway together with a low-rank LoRA adapter \cite{hu2021loralowrankadaptationlarge} on the weights of $\Phi$, keeping the remaining Trellis.2 parameters frozen. Training is performed on synthetic scene data, supervising chunk generation against ground-truth chunk latents extracted from the synthetic scenes. Further details are specified in Appendix \ref{app:setup}.

\subsection{Scene reconstruction at test time}
\label{sec:pipeline}

At test time, given an unordered set of RGB images \( \{I_n\}_{n=1}^N \) of an unseen scene, we produce a scene-level PBR mesh \( \mathcal{M} \). 

\paragraph{Scene calibration and chunking.}
Since the input images are unposed at inference time, we first run structure-from-motion (COLMAP~\cite{schonbergerStructurefromMotionRevisited2016}) to recover the camera intrinsics \( \{K_n\} \), extrinsics \( \{T_n\} \), and a sparse point cloud \( P \subset \mathbb{R}^3 \) of the scene. We apply statistical and radius-based outlier filtering to \( P \) and estimate the scene's spatial extent \( \mathcal{B} \subset \mathbb{R}^3 \) from the filtered points using robust percentile-based bounds.
Given \( \mathcal{B} \), we partition the scene volume into a set of chunks \( \mathcal{C} = \{c_1, \dots, c_K\} \), each occupying a fixed-size cube \( V_c \) in its own canonical frame with a translation \( t_c \in \mathbb{R}^3 \) placing it in the world frame. Neighboring chunks overlap by a prescribed minimum margin \( m \), providing the regions across which chunks exchange information during joint generation. 

\paragraph{Global 3D conditioning.}
Rather than computing the conditioning grids \( G^{(c)} \) independently per chunk, we compute a global conditioning grid \( G \) once over the full scene volume and extract per-chunk conditions \( G^{(c)} \) as crops. Concretely, we lift each encoded image \( F_n \) into a scene-sized voxel grid via the per-view projection of Section~\ref{sec:prior}, and aggregate across views to obtain \( G \). For occupancy generation, \( G\) is dense at the resolution of the occupancy latents; for shape and texture generation, which operate on higher-resolution sparse latents defined by the predicted occupancy, the lifting and aggregation are also performed on the corresponding sparse high-resolution voxel structure. The per-chunk conditions \( G^{(c)} \) are then obtained by cropping \( G \) to each \( V_c \).

\paragraph{Joint chunk generation.}
All chunks are generated jointly by $\Phi$ in a single flow-matching trajectory following a MultiDiffusion-style~\cite{bartal2023multidiffusionfusingdiffusionpaths} scheme. We maintain a global noisy latent grid \( z_t \) covering the full scene volume. At each step \( t \), for each chunk \( c \in \mathcal{C} \) we extract its corresponding latent crop \( \smash{z^{(c)}_t} \) and apply the chunk-wise denoiser to obtain a per-chunk prediction \( \smash{\hat{z}_{t-1}^{(c)}} \). The per-chunk predictions are merged into the next global latent \( z_{t-1} \) by averaging in overlap regions:
\begin{equation}
z_{t-1}(x) = \frac{1}{\sum_{c \in \mathcal{C}} M_c(x)}
\sum_{c \in \mathcal{C}} M_c(x)\, \hat{z}_{t-1}^{(c)}(x),
\end{equation}
where \( x \) indexes a voxel in the global scene grid and \( M_c(x) \in \{0,1\} \) indicates whether \( x\) lies within \( V_c \). This aggregation enforces consistency across chunk boundaries throughout the generation trajectory. For shape and texture generation, we additionally apply a boundary-sensitive variant in which chunk-boundary voxels do not contribute to the aggregation but are still updated by it; we find this improves seam coherence visually. After the final step, the global latent grid \( z_0 \) is decoded by the respective Trellis.2 decoders into the final scene mesh \( \mathcal{M} \) with PBR materials.

\section{Experiments}

\paragraph{Datasets.}
We train on chunks extracted from synthetic indoor scene data.
Our primary training dataset is SAGE-10k \cite{xia2026sagescalableagentic3d}, a set of synthetic indoor scenes with PBR materials and objects generated by Trellis \cite{xiang2024structured}. 
While SAGE-10k provides a wide variety of single rooms, it does not contain multi-room layouts, windows, or door openings, all of which are important for our model to perform on real-world scenes.
To expose our model to these structural elements, we additionally include a subset of scenes from 3D-FRONT \cite{fu20203dfront} for occupancy generation training.
See Appendix \ref{app:setup} for details.

\paragraph{Evaluation.}
We evaluate on unseen scenes from two datasets: 3D-FRONT \cite{fu20203dfront} and ScanNet++~\cite{yeshwanthliu2023scannetpp}, to assess performance on synthetic and out-of-domain real-world data.
For both settings, we evaluate 25 scenes with 8 input views each.
Additionally, we assume a set of sparse points and the camera poses to be given.
For 3D-FRONT, we use single-room scenes with ground-truth poses, and sample 10k points from backprojected training-view depth maps to obtain points that respect visibility constraints.
For ScanNet++, we use the provided COLMAP \cite{schonbergerStructurefromMotionRevisited2016} outputs.

\paragraph{Metrics.} 
We evaluate reconstructed meshes in both 2D and 3D. In 2D, we report geometric errors (MAE, RMSE, AbsRel, SqRel, angular normal error), perceptual/semantic metrics (LPIPS, CLIP), and completeness over valid pixels only; in 3D, we measure alignment and coverage using Chamfer distance, F-score (10 cm), and normal consistency (thresholded at 20 cm), restricted to observed regions.
Details are specified in Appendix~\ref{app:metrics}.

\paragraph{Baselines.}
We compare against five reconstruction methods that span dominant paradigms in the literature.
2D Gaussian Splatting~(2DGS)~\cite{huang20252dgaussiansplattinggeometrically} performs prior-free per-scene optimization.
MonoSDF~\cite{yu2022monosdfexploringmonoculargeometric} uses monocular geometric priors to help guide per-scene optimization.
Depth~Anything~3~(DA3)~\cite{lin2025depth3recoveringvisual} is a feed-forward monocular depth foundation model; we fuse its predicted depths into 3D meshes using TSDF fusion.
FineRecon~\cite{stier2023finerecondepthawarefeedforwardnetwork} performs 3D refinement of fused monocular predictions, for which we use DA3 as the underlying depth foundation model.
Murre~\cite{guo2025multiviewreconstructionsfmguidedmonocular} uses diffusion-based depth priors with 3D conditioning.
Please refer to Appendix~\ref{app:baselines} for additional details of the baselines.

\subsection{Reconstruction Results}
\begin{table}[htb]
\centering
\caption{\textbf{2D baseline comparisons on real-world data.} We evaluate 2D depth and normal metrics on 25 scenes from ScanNet++.}
\label{tab:scannetpp_metrics_2d}
\setlength{\tabcolsep}{4pt}
\begin{tabular}{llrrrrrr}
\toprule
 &  & 2DGS & MonoSDF & DA3 & FineRecon & Murre & Ours \\
\midrule
\multirow{6}{*}{Depth} & MAE (m) $\downarrow$ & 0.2599 & 0.2496 & 0.1235 & \cellcolor{red!25}\textbf{0.0939} & \cellcolor{yellow!25}0.1072 & \cellcolor{orange!25}0.1024 \\
 & RMSE (m) $\downarrow$ & 0.4535 & 0.3812 & \cellcolor{yellow!25}0.2428 & \cellcolor{orange!25}0.2147 & 0.2524 & \cellcolor{red!25}\textbf{0.2100} \\
 & AbsRel $\downarrow$ & 0.2296 & 0.3379 & 0.1627 & \cellcolor{yellow!25}0.1186 & \cellcolor{orange!25}0.1177 & \cellcolor{red!25}\textbf{0.1138} \\
 & SqRel $\downarrow$ & 0.1763 & 0.3251 & 0.1000 & \cellcolor{yellow!25}0.0853 & \cellcolor{orange!25}0.0846 & \cellcolor{red!25}\textbf{0.0532} \\
 & LPIPS $\downarrow$ & 0.3702 & \cellcolor{orange!25}0.1143 & 0.2966 & \cellcolor{yellow!25}0.1716 & 0.3060 & \cellcolor{red!25}\textbf{0.0872} \\
 & CLIP score $\uparrow$ & 0.8939 & \cellcolor{orange!25}0.9718 & 0.9258 & \cellcolor{yellow!25}0.9404 & 0.9137 & \cellcolor{red!25}\textbf{0.9803} \\
\midrule
\multirow{3}{*}{Normal} & Angular err. ($^\circ$) $\downarrow$ & 35.9195 & \cellcolor{yellow!25}20.2113 & 26.2189 & \cellcolor{orange!25}19.1835 & 23.2825 & \cellcolor{red!25}\textbf{16.0179} \\
 & LPIPS $\downarrow$ & 0.7047 & \cellcolor{orange!25}0.4201 & 0.6562 & \cellcolor{yellow!25}0.4888 & 0.6141 & \cellcolor{red!25}\textbf{0.3411} \\
 & CLIP score $\uparrow$ & 0.8679 & \cellcolor{orange!25}0.9158 & \cellcolor{yellow!25}0.8995 & 0.8920 & 0.8844 & \cellcolor{red!25}\textbf{0.9439} \\
\midrule
 & Completeness $\uparrow$ & 0.7202 & \cellcolor{orange!25}0.9716 & 0.8404 & \cellcolor{yellow!25}0.8707 & 0.7300 & \cellcolor{red!25}\textbf{0.9904} \\
\bottomrule
\end{tabular}%
\end{table}
\begin{table}[htb]
\centering
\caption{\textbf{3D baseline comparisons on real-world data.} We evaluate 3D reconstruction metrics on 25 scenes from ScanNet++.}
\label{tab:scannetpp_metrics_3d}
\setlength{\tabcolsep}{4pt}
\begin{tabular}{lrrrrrr}
\toprule
 & 2DGS & MonoSDF & DA3 & FineRecon & Murre & Ours \\
\midrule
Chamfer (m) $\downarrow$ & 0.2181 & 0.1250 & 0.0998 & \cellcolor{orange!25}0.0819 & \cellcolor{yellow!25}0.0978 & \cellcolor{red!25}\textbf{0.0688} \\
F-score @10cm $\uparrow$ & 0.4870 & 0.6317 & 0.6905 & \cellcolor{orange!25}0.7694 & \cellcolor{yellow!25}0.7263 & \cellcolor{red!25}\textbf{0.7771} \\
Normal consistency $\uparrow$ & 0.4903 & 0.7039 & \cellcolor{yellow!25}0.7140 & \cellcolor{orange!25}0.7851 & 0.7125 & \cellcolor{red!25}\textbf{0.7860} \\
\bottomrule
\end{tabular}%
\end{table}

\paragraph{Real-world scene reconstruction.}
In Table \ref{tab:scannetpp_metrics_2d} and Table \ref{tab:scannetpp_metrics_3d}, we evaluate the performance on ScanNet++ \cite{yeshwanthliu2023scannetpp}.
Despite training only on synthetic data, our method achieves the strongest reconstruction quality on this entirely unseen real-world dataset across both 2D and 3D metrics, with substantially better perceptual and semantic alignment with the ground-truth laser scans on depth and normals and the highest completeness of all methods evaluated.

\begin{figure}[h]
    \centering
    \includegraphics[width=1\linewidth]{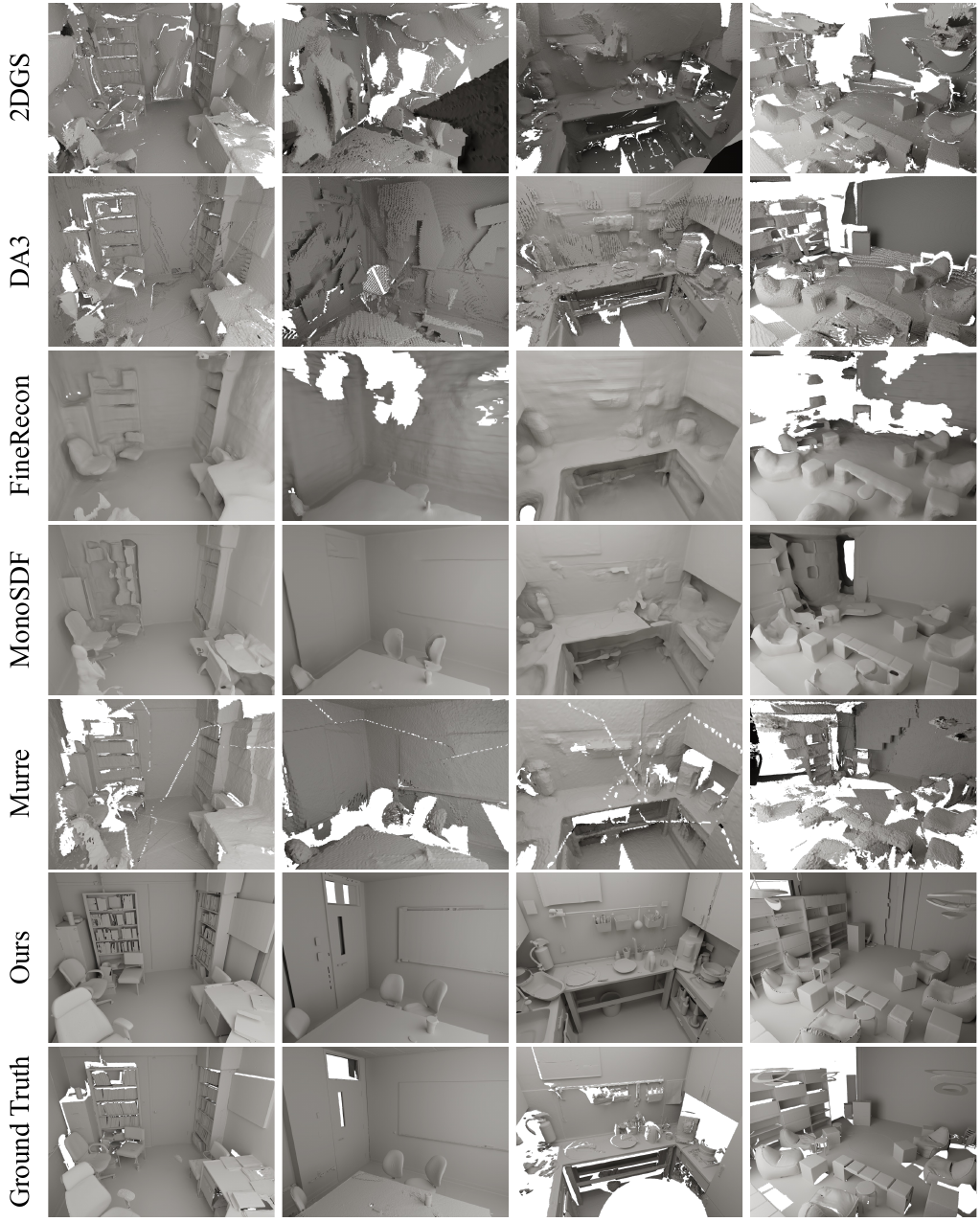}
    \caption{\textbf{Real-world comparisons.} Reconstruction performance on four evaluation scenes from ScanNet++. Compared to the baselines our results are more complete and reconstruct finer details.}
    \label{fig:qualitative_comp}
\end{figure}

Figure \ref{fig:qualitative_comp} qualitatively compares the performance of our method against the baselines on ScanNet++ using 8 input images. While the baselines produce noisy (2DGS, DA3), oversmooth (FineRecon, MonoSDF) surfaces for challenging areas and are incomplete in occluded and unobserved areas (2DGS, DA3, FineRecon, Murre), our approach yields complete and high-fidelity reconstructions.

\paragraph{Synthetic scene reconstruction.}
Tables~\ref{tab:vfront_metrics_2d} and~\ref{tab:vfront_metrics_3d} report 2D and 3D metrics on 3D-FRONT~\cite{fu20203dfront}.
While our occupancy stage was fine-tuned on a small subset of 3D-FRONT scenes, evaluation is performed exclusively on held-out scenes. Our method again achieves the strongest overall performance across both 2D and 3D metrics, indicating that the recovered geometry not only minimizes error but also matches the structural characteristics of the ground truth, avoiding both oversmoothing and high-frequency artifacts.

\begin{table}[htb]
\centering
\caption{\textbf{2D baseline comparisons on synthetic data.} We evaluate 2D depth and normal metrics on 25 scenes from 3D-FRONT.}
\label{tab:vfront_metrics_2d}
\setlength{\tabcolsep}{4pt}
\begin{tabular}{llrrrrrr}
\toprule
 &  & 2DGS & MonoSDF & DA3 & FineRecon & Murre & Ours \\
\midrule
\multirow{6}{*}{Depth} & MAE (m) $\downarrow$ & 0.4542 & 0.8134 & \cellcolor{orange!25}0.0957 & 0.1334 & \cellcolor{red!25}\textbf{0.0933} & \cellcolor{yellow!25}0.1131 \\
 & RMSE (m) $\downarrow$ & 0.8475 & 1.0947 & \cellcolor{yellow!25}0.3712 & 0.4024 & \cellcolor{orange!25}0.3396 & \cellcolor{red!25}\textbf{0.3200} \\
 & AbsRel $\downarrow$ & 0.2191 & 0.5546 & \cellcolor{orange!25}0.0686 & 0.1121 & \cellcolor{yellow!25}0.0729 & \cellcolor{red!25}\textbf{0.0609} \\
 & SqRel $\downarrow$ & 0.3629 & 1.2379 & \cellcolor{yellow!25}0.1259 & 0.1977 & \cellcolor{orange!25}0.1205 & \cellcolor{red!25}\textbf{0.0587} \\
 & LPIPS $\downarrow$ & 0.5693 & \cellcolor{orange!25}0.2538 & 0.4622 & 0.4289 & \cellcolor{yellow!25}0.3892 & \cellcolor{red!25}\textbf{0.1270} \\
 & CLIP score $\uparrow$ & 0.8581 & \cellcolor{orange!25}0.9349 & 0.8819 & 0.8677 & \cellcolor{yellow!25}0.8907 & \cellcolor{red!25}\textbf{0.9565} \\
\midrule
\multirow{3}{*}{Normal} & Angular err. ($^\circ$) $\downarrow$ & 38.1341 & 24.6739 & \cellcolor{yellow!25}16.0993 & \cellcolor{orange!25}15.4063 & 16.9646 & \cellcolor{red!25}\textbf{10.7490} \\
 & LPIPS $\downarrow$ & 0.8278 & \cellcolor{orange!25}0.4580 & 0.6256 & \cellcolor{yellow!25}0.5838 & 0.6081 & \cellcolor{red!25}\textbf{0.2183} \\
 & CLIP score $\uparrow$ & 0.8155 & \cellcolor{yellow!25}0.8752 & \cellcolor{orange!25}0.8842 & 0.8531 & 0.8721 & \cellcolor{red!25}\textbf{0.9571} \\
\midrule
 & Completeness $\uparrow$ & 0.3924 & \cellcolor{orange!25}0.9002 & 0.6424 & 0.5627 & \cellcolor{yellow!25}0.7484 & \cellcolor{red!25}\textbf{0.9871} \\
\bottomrule
\end{tabular}%
\end{table}

\begin{table}[htb]
\centering
\caption{\textbf{3D baseline comparisons on synthetic data.} We evaluate 3D reconstruction metrics on 25 scenes from 3D-FRONT.}
\label{tab:vfront_metrics_3d}
\setlength{\tabcolsep}{4pt}
\begin{tabular}{lrrrrrr}
\toprule
 & 2DGS & MonoSDF & DA3 & FineRecon & Murre & Ours \\
\midrule
Chamfer (m) $\downarrow$ & 0.5924 & 0.3387 & \cellcolor{yellow!25}0.2087 & 0.2227 & \cellcolor{orange!25}0.1584 & \cellcolor{red!25}\textbf{0.0638} \\
F-score @10cm $\uparrow$ & 0.1766 & 0.3463 & \cellcolor{yellow!25}0.5501 & 0.5129 & \cellcolor{orange!25}0.6014 & \cellcolor{red!25}\textbf{0.8655} \\
Normal consistency $\uparrow$ & 0.3481 & 0.4264 & \cellcolor{yellow!25}0.6464 & 0.6244 & \cellcolor{orange!25}0.6624 & \cellcolor{red!25}\textbf{0.8199} \\
\bottomrule
\end{tabular}%
\end{table}

\subsection{Ablations}
\label{sec:ablations}
We ablate the projection-based 3D conditioning pathway and the number of input views.
The ablations are performed on 25 chunks drawn from 25 distinct SAGE-10k \cite{xia2026sagescalableagentic3d} scenes not seen during training.
Qualitative results are provided in Appendix~\ref{app:results}.

\paragraph{Effectiveness of the 3D contitioning path.}
We compare three variants: 
(i) the vanilla pretrained Trellis.2 model; 
(ii) Trellis.2 fine-tuned on our scene data; 
(iii) our full method. 
As shown in Table~\ref{tab:sage_metrics_3d}, the vanilla Trellis.2 performs poorly on scene chunks since it was trained for object-level generation with full visibility in the conditioning image. 
Fine-tuning Trellis.2 on scene data without the 3D condition produces plausible scene fragments but fails to place them in the correct pose. 
Adding our 3D conditioning pathway resolves this, aligning the generated geometry with the ground-truth pose. 
This confirms that the 3D condition is what enables pose-controlled chunk generation.

\paragraph{Effect of the number of input views.}
We further evaluate the sensitivity of our method to the number of conditioning views per chunk. 
We report results in Table~\ref{tab:sage_metrics_3d}.
Notably, our 3D conditioning enables pose-correct chunk generation from already a single input image.
Performance improves with additional views, as more of the chunk is directly observed. 

\begin{table}[htb]
\centering
\caption{\textbf{3D ablation.} We evaluate the efficacy of our 3D conditioning on 25 scenes from SAGE-10k.}
\label{tab:sage_metrics_3d}
\setlength{\tabcolsep}{4pt}
\resizebox{\linewidth}{!}{%
\begin{tabular}{lrrrrrr}
\toprule
 & 1img vanilla & 1img no3Dcond & 1img & 2img & 4img & 8img \\
\midrule
Chamfer (m) $\downarrow$ & 0.3450 & 0.2578 & 0.1345 & \cellcolor{yellow!25}0.0812 & \cellcolor{orange!25}0.0420 & \cellcolor{red!25}\textbf{0.0291} \\
F-score @10cm $\uparrow$ & 0.2067 & 0.3666 & 0.6312 & \cellcolor{yellow!25}0.7608 & \cellcolor{orange!25}0.9070 & \cellcolor{red!25}\textbf{0.9683} \\
Normal consistency (20cm cap) $\uparrow$ & 0.2341 & 0.3549 & 0.6349 & \cellcolor{yellow!25}0.7688 & \cellcolor{orange!25}0.8411 & \cellcolor{red!25}\textbf{0.8804} \\
\bottomrule
\end{tabular}%
}
\end{table}

\subsection{PBR Texture and Relighting}

\begin{figure}[htb]
    \centering
    \includegraphics[width=1\linewidth]{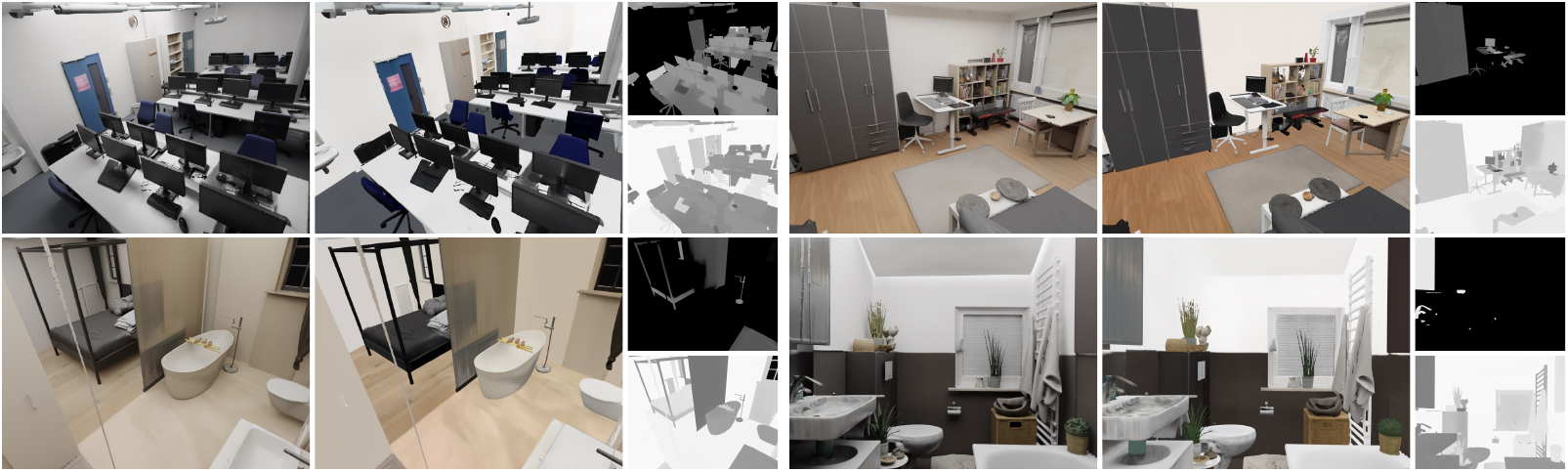}
    \caption{\textbf{PBR results.} Qualitative results on four reconstructed scenes from ScanNet++: lit scene (left), albedo (middle), metallic and roughness (right).}
    \label{fig:pbr}
\end{figure}

\begin{figure}[htb]
    \centering
    \includegraphics[width=1\linewidth]{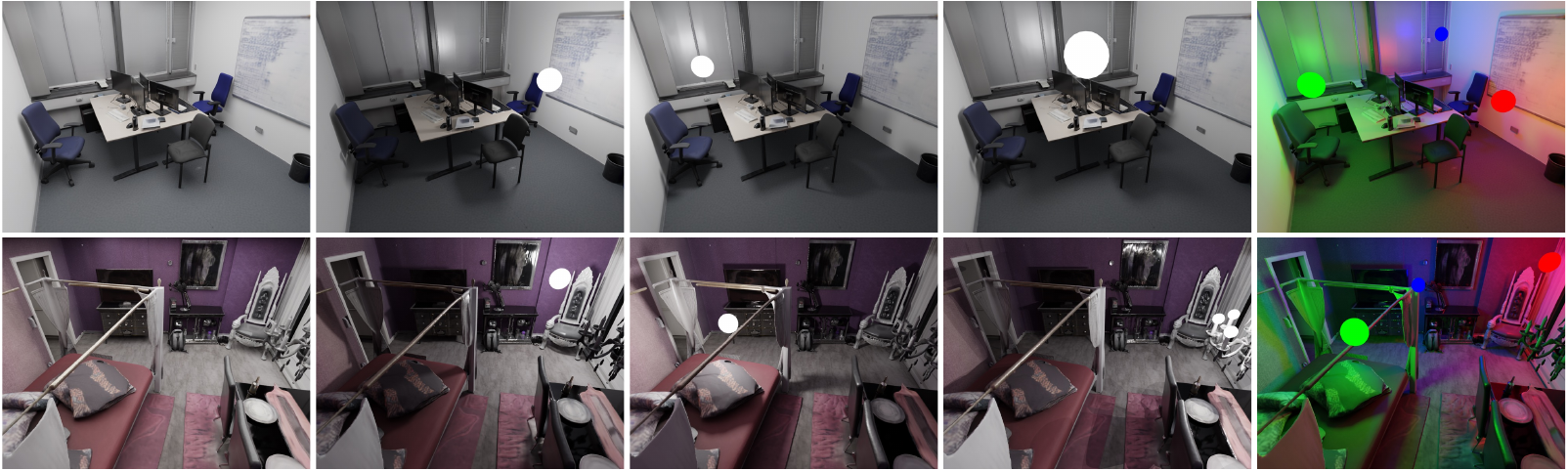}
    \caption{\textbf{Relighting results.} Varying lighting configurations for scenes reconstructed from ScanNet++. Further visualizations are provided in Appendix~\ref{app:results}.}
    \label{fig:relighting}
\end{figure}

Our method produces scene-level meshes with physically-based material properties (albedo, metallic, roughness), directly importable into standard rendering engines and relightable under arbitrary illumination without any per-scene optimization. 
Figure~\ref{fig:pbr} shows the predicted material channels alongside the lit reconstruction, and Figure~\ref{fig:relighting} shows relighting on real-world ScanNet++ scenes, with recovered albedo and reflectance responding plausibly in each case.
Since our texture stage is fine-tuned exclusively on SAGE-10k, where PBR materials are themselves VLM-predicted rather than ground-truth measured, the recovered textures are visually plausible but do not match dedicated SVBRDF-estimation methods in absolute material accuracy.

\subsection{Limitations}
\label{sec:limitations}
While our method achieves strong reconstruction quality across a wide range of indoor scenes, several limitations remain.
Reconstructions of non-Lambertian surfaces, such as glass and mirrors, are less reliable, as such materials are underrepresented in SAGE-10k. 
Our chunk partitioning is currently designed for indoor scene settings with vertical extents up to roughly 5m. 
Extending beyond this would require adaptive chunking based on the full spatial extent.
Finally, our incorporation of a strong generative prior may lead to reconstructions occasionally hallucinating content in regions where input views provide weak evidence. In practice, we find quantitatively and qualitatively that the strong scene prior nonetheless yields substantial gains over pure reconstruction baselines, with the prior's tendency to fill in plausible structure improving completeness and surface fidelity far more than occasional hallucinations detract from accuracy.


\section{Conclusion}
\label{sec:conclusion}

We presented a method for reconstructing scene-level PBR meshes from posed multi-view images by lifting an object-level 3D generative prior to scene scale. 
Our key idea is to formulate scene reconstruction as the joint generation of overlapping spatial chunks, conditioned on the input views through a projection-based 3D conditioning pathway that injects multi-view image features into the generative model in a spatially grounded manner. 
Combined with parameter-efficient LoRA fine-tuning on synthetic indoor data, this enables faithful, editable PBR mesh reconstructions that inherit the fidelity and completeness of modern object-level generative priors while extending naturally to scene scale. 
We see this work as a step toward closing the gap between current reconstruction methods and the quality required for downstream graphics, simulation, and embodied AI pipelines.

\section*{Acknowledgements}

Katharina Schmid is a doctoral fellow at the Munich Data Science Institute (MDSI), and Nicolas von Lützow was supported by the MDSI focus topic \textit{Understanding Existing Structures in Building Planning} (USP).
Angela Dai was supported by the ERC Starting Grant SpatialSem (101076253).
This work was further supported by the ERC Consolidator Grant \textit{Gen3D} (101171131), by compute resources from the Jülich Supercomputing Center under project \textit{MeshFoundation}, and by the Computing Systems Lab, part of the Huawei Technologies Switzerland AG.

\bibliographystyle{unsrtnat}
\bibliography{bibliography}


\appendix
\newpage
\section{Experimental Setup}\label{app:setup}

\paragraph{Implementation Details.}
We build on Trellis.2 \cite{xiang2025nativecompactstructuredlatents} at resolution 512.
Consequently, for the occupancy generation, we use a dense latent grid of resolution 16, and for the shape and texture generation, a sparse latent grid of resolution 32 per chunk.
We apply LoRA adapters \cite{hu2021loralowrankadaptationlarge} to all attention layers with rank 8 and scaling 16.
The 3D condition is injected into each of the 30 DiT blocks through a per-block zero-initialized projection that is residually added to the block's current representation.
The aggregation network consists of two parallel MLPs sharing the same per-view input: a three-layer feature MLP and a two-layer scalar MLP that produces per-view aggregation logits, which are softmaxed across views to form attention-style aggregation weights.

\paragraph{Training Details.}
We train with a global batch size of 32. 
We use AdamW \cite{loshchilov2019decoupledweightdecayregularization} with a learning rate of \( 1 \times 10^{-4} \) for newly introduced layers and \( 3 \times 10^{-5} \) for LoRA parameters.
The learning rate decay is set to \( 0.01 \) for new parameters and \( 0.0 \) for LoRA parameters.
During training, the number of conditioning views per chunk is uniformly sampled from 1 to 16.
We train with classifier-free guidance \cite{ho2022classifierfreediffusionguidance}, dropping the condition with a probability of \( 0.1 \).

\paragraph{Inference Details.}
The chunk size is estimated based on the sparse point cloud from SfM.
Specifically, we choose the chunk size as \( 1.11 \times \text{scene height}\).
Chunks are placed such that neighboring chunks overlap by a prescribed minimum margin \( m = 0.25\).
For shape and texture generation, we apply a boundary-sensitive variant of the voxel feature synchronization.
Specifically, the boundary-sensitive variant modifies the MultiDiffusion \cite{bartal2023multidiffusionfusingdiffusionpaths} overlap aggregation by excluding voxels in the outermost \( b = 1 \) rows of each chunk's x/y faces from contributing to the cross-chunk mean, since these boundary regions are systematically biased by the network's lack of receptive-field context beyond the chunk extent. 
Every voxel, including boundaries, still receives the resulting interior-only average wherever at least one chunk's interior covers it, so seams are healed by trustworthy interior predictions while voxels with no interior coverage fall back to their own chunk's estimate.
Following the original Trellis.2 implementation, we use 12 flow-matching steps for each model at inference time.

\paragraph{Dataset Curation.}
We use a subset of 5,000 rooms from SAGE-10k \cite{xia2026sagescalableagentic3d}. 
For each room, we randomly place light sources, sample random camera poses with a per-room sampled focal length, and render the resulting RGB images. 
For each room, we extract three randomly-rotated training chunks, yielding 15,000 chunk-level training examples in total.
Additionally, we include a subset of scenes from 3D-FRONT \cite{fu20203dfront}: 1,200 houses, 3 chunks per house, used only for the occupancy stage.

\section{Metrics Details}\label{app:metrics}
We evaluate reconstruction quality both on unseen 2D test views and directly on the extracted 3D meshes.

In the 2D setting, we render depth and normal maps from the reconstructed meshes at held-out viewpoints.
For depth, we report the mean absolute error (MAE) and root mean squared error (RMSE) in meters, together with their relative counterparts AbsRel and SqRel, measuring depth accuracy.
For normals, we report the angular normal error in degrees to assess view-space surface orientations.
We additionally measure perceptual and semantic agreement with the ground-truth views using LPIPS~\cite{zhang2018perceptual} with an AlexNet backbone~\cite{NIPS2012_c399862d} and CLIP score~\cite{radford2021learningtransferablevisualmodels} on both depth and normal renderings.
Finally, we report prediction completeness.
All metrics are computed only over pixels where ground-truth views are defined, excluding empty or unobserved regions and avoiding ambiguities caused by windows and mirrors.

In the 3D setting, we evaluate the Chamfer distance to measure average 3D alignment and the F-score to measure areas within a reasonable tolerance.
Additionally, we evaluate normal consistency between nearest-neighbor surface points to measure local surface orientation.

\paragraph{Masking.}
For 2D evaluation, views are rendered at \(876{\times}584\) resolution on ScanNet++ and at \(512{\times}512\) on 3D-FRONT. 
Pixel-wise depth and normal metrics are computed only over the intersection of valid ground-truth and predicted pixels, such that they measure reconstruction quality where both meshes produce a hit. 
Missing predictions are instead captured by the separate completeness metric. 
For perceptual and image-level metrics such as LPIPS, FID/KID, and CLIP score, we apply the GT mask to the predicted images to avoid punishing results in regions that are undefined for the GT laser scan. 

For 3D evaluation, metrics are computed inside a per-scene observation envelope rather than over the full extent of the predicted mesh. 
On ScanNet++, this envelope is derived from the full-scanner fusion at \(1.1\)~cm voxel resolution and dilated by \(15\)~cm, yielding a permissive region around observed surfaces that gates evaluation without acting as a tight accuracy threshold. 
Predicted meshes are clipped to this envelope before sampling, removing geometry outside the scanned region. 
For 3D-FRONT, we instead use the ground-truth mesh axis-aligned bounding box inflated by \(20\%\).

\paragraph{2D Metrics.}
We aggregate image-space metrics over all valid pixels pooled across all test frames.
Let \(d_{f,i}\) and \(n_{f,i}\) denote the predicted depth and view-space unit normal at pixel \(i\) in frame \(f\), with \(d^*_{f,i}\) and \(n^*_{f,i}\) denoting the corresponding ground truth.
The set \(\mathcal{I}\) contains all frame-pixel pairs where both predicted and ground-truth depth are valid; depth and normal errors are averaged over \(\mathcal{I}\).
\begin{align}
    \text{MAE:}\quad
    &\frac{1}{|\mathcal{I}|}
    \sum_{(f,i) \in \mathcal{I}}
    \left| d_{f,i} - d^*_{f,i} \right|,
    \\
    \text{RMSE:}\quad
    &\sqrt{
    \frac{1}{|\mathcal{I}|}
    \sum_{(f,i) \in \mathcal{I}}
    \left( d_{f,i} - d^*_{f,i} \right)^2
    },
    \\
    \text{AbsRel:}\quad
    &\frac{1}{|\mathcal{I}|}
    \sum_{(f,i) \in \mathcal{I}}
    \frac{\left| d_{f,i} - d^*_{f,i} \right|}{d^*_{f,i}},
    \\
    \text{SqRel:}\quad
    &\frac{1}{|\mathcal{I}|}
    \sum_{(f,i) \in \mathcal{I}}
    \frac{\left( d_{f,i} - d^*_{f,i} \right)^2}{d^*_{f,i}},
    \\
    \text{Normal error:}\quad
    &\frac{1}{|\mathcal{I}|}
    \sum_{(f,i) \in \mathcal{I}}
    \arccos\!\left(
        n_{f,i}^{\top} n^*_{f,i}
    \right)
    \cdot
    \frac{180}{\pi}.
\end{align}
For LPIPS and CLIP score, we compare saved ground-truth and predicted renderings per held-out test view and average the resulting scores over all views.
Let \(\mathcal{F}\) denote the set of evaluated test frames, \(m\) the evaluated modality, i.e., depth or normal, and \(I^m_f\) and \(I^{m,*}_f\) the predicted and ground-truth rendered images for modality \(m\) at frame \(f\).
\begin{align}
    \text{LPIPS}_{m}:\quad
    &\frac{1}{|\mathcal{F}|}
    \sum_{f \in \mathcal{F}}
    \mathrm{LPIPS}\!\left(
        I^m_f,
        I^{m,*}_f
    \right),
    \\
    \text{CLIP}_{m}:\quad
    &\frac{1}{|\mathcal{F}|}
    \sum_{f \in \mathcal{F}}
    \frac{
        \phi_{\mathrm{CLIP}}(I^m_f)^\top
        \phi_{\mathrm{CLIP}}(I^{m,*}_f)
    }{
        \left\|\phi_{\mathrm{CLIP}}(I^m_f)\right\|_2
        \left\|\phi_{\mathrm{CLIP}}(I^{m,*}_f)\right\|_2
    }.
\end{align}

\paragraph{3D Metrics.}
For 3D evaluation, we uniformly sample \(200\)k points from both the predicted and ground-truth meshes and compute nearest-neighbor distances in both directions.

Chamfer distance is reported as the mean of the predicted-to-ground-truth and ground-truth-to-predicted nearest-neighbor distances in meters.

For F-score@10cm, we compute precision as the fraction of predicted samples within \(0.1\)~m of the ground-truth surface and recall as the fraction of ground-truth samples within \(0.1\)~m of the predicted surface, and then report their mean.

Normal consistency is computed symmetrically over the same nearest-neighbor correspondences as the average absolute normal dot product in both directions.
To avoid rewarding geometrically unrelated but similarly oriented surfaces, normal agreement is set to zero for nearest-neighbor pairs farther than \(0.2\)~m.

\section{Further Results}\label{app:results}

\begin{figure}[h]
    \centering
    \includegraphics[width=1\linewidth]{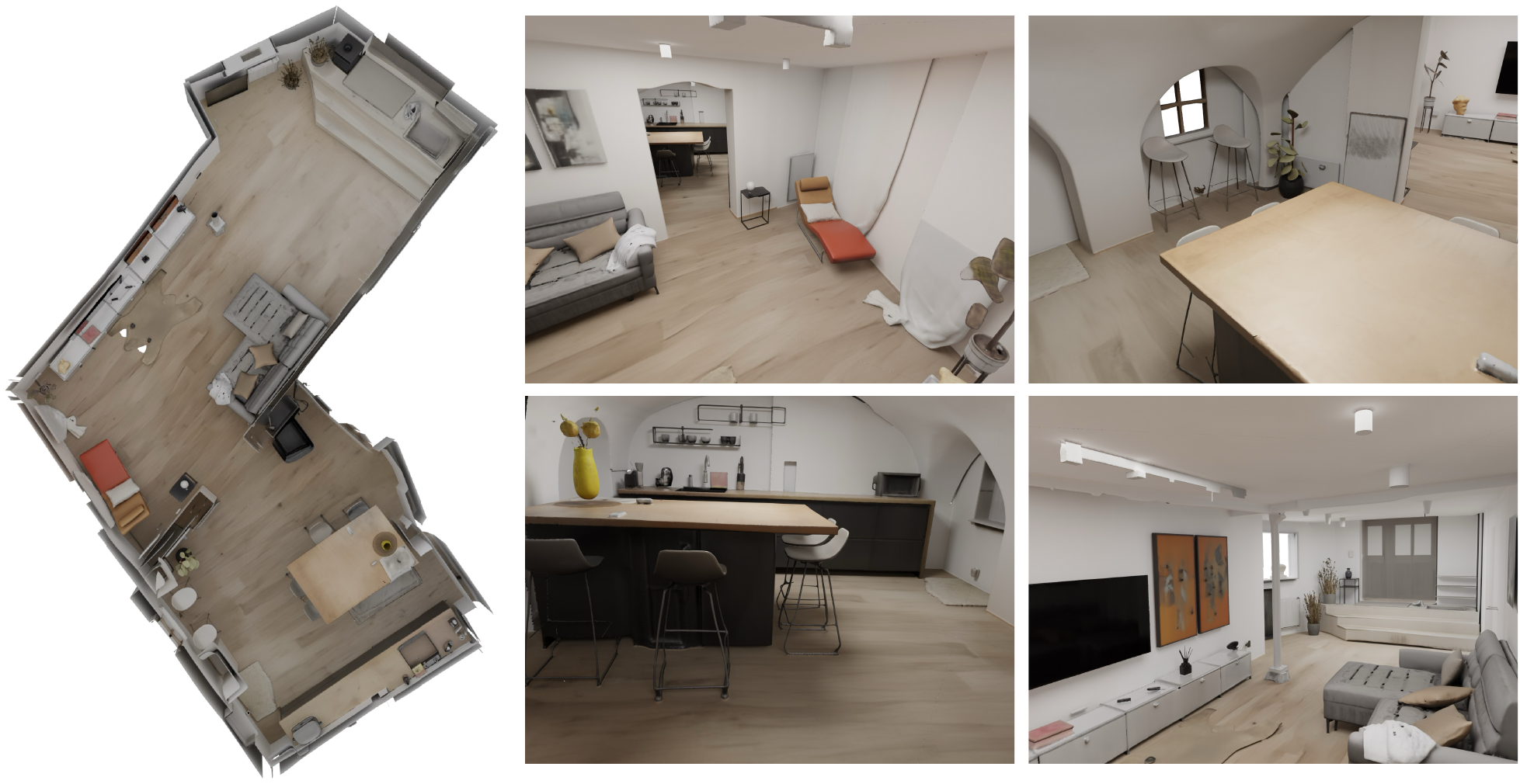}
    \caption{\textbf{Large Scene Generation.} Top-down view (left) and multiple close-ups (right).}
    \label{fig:large_scene}
\end{figure}

\paragraph{Large Scene Generation}
As visualized in Figure \ref{fig:large_scene}, our method yields high-fidelity results on large 3D indoor scenes from ScanNet++.

\begin{figure}[h]
    \centering
    \includegraphics[width=1\linewidth]{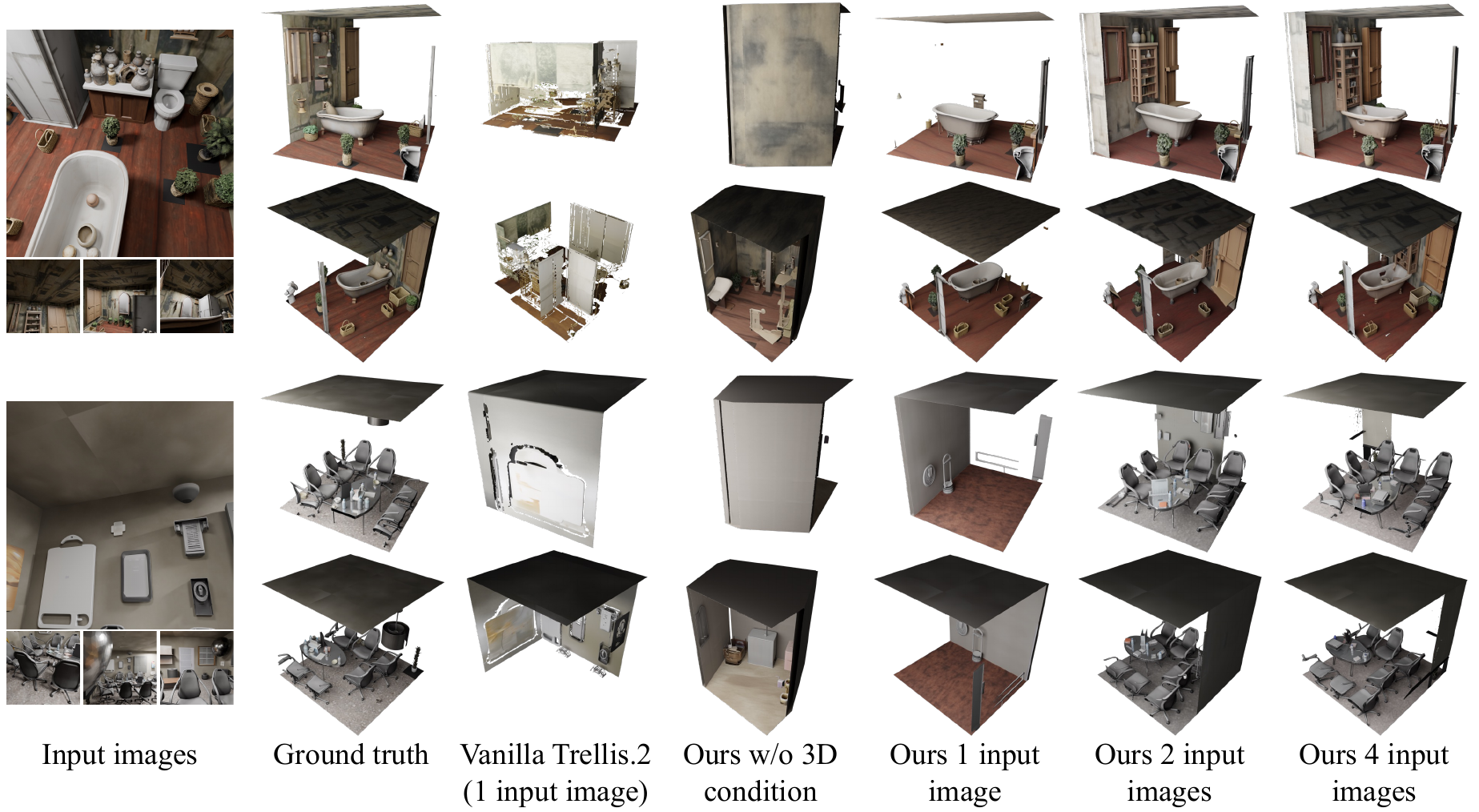}
    \caption{\textbf{Ablation results.} Ablation study on SAGE-10k chunks not seen during training. Our projection-based 3D conditioning effectively enables pose-correct chunk generation from as few as a single input image. Reconstruction quality improves with additional views.}
    \label{fig:ablation}
\end{figure}

\paragraph{Ablation}
Figure \ref{fig:ablation} visualizes the qualitative results of our ablation study as described in Section~\ref{sec:ablations}. 
Vanilla Trellis.2 fails on scene chunks due to its object-level training, and while fine-tuning on scenes improves local plausibility, it cannot recover correct pose without our 3D conditioning. 
Our 3D conditioning enables pose-consistent generation, even from a single view, and performance further improves as more input views provide increased scene coverage.

\begin{figure}[h]
    \centering
    \includegraphics[width=1\linewidth]{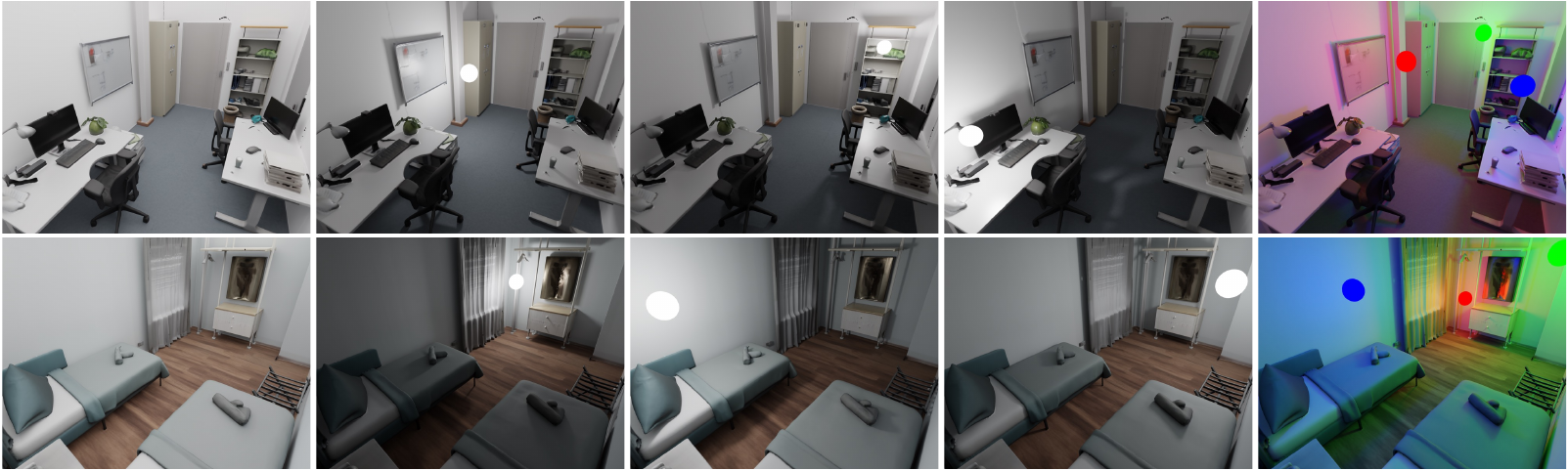}
    \caption{\textbf{Additional relighting results.} Varying lighting configurations for scenes reconstructed from ScanNet++.}
    \label{fig:relighting_2}
\end{figure}

\paragraph{Relighting}
Additional relighting results are provided in Figure~\ref{fig:relighting_2}.

\section{Baseline Implementation Details}\label{app:baselines} 
We evaluate all baselines in the same posed multi-view setting as our method.
Whenever possible, we use the exact released configurations and only make changes required to adapt the methods to our evaluation protocol.
All such changes fix issues and make the methods applicable under our sparse-view evaluation; none of them weaken the baselines.

\paragraph{2D Gaussian Splatting \cite{huang20252dgaussiansplattinggeometrically}.}
We use the official 2D Gaussian Splatting implementation and optimize scenes independently.
The hyperparameters follow the authors' Tanks-and-Temples large-scene preset, adjusting only the TSDF fusion to keep the top \(50\) clusters by triangle count, rather than just one, to avoid removing correct geometry from unconnected regions.

\paragraph{Depth-Anything-3 \cite{lin2025depth3recoveringvisual}.}
We run the pretrained \emph{DA3NESTED-GIANT-LARGE-1.1} model to return RGB-D predictions jointly for all input images in a single forward pass.
During TSDF fusion, we use a voxel size of \(0.01\)~m, truncation distance \(0.04\)~m, and maximum depth \(4.5\)~m.
We again keep the top \(50\) clusters of the resulting mesh.
The predicted depths are also reused as input to FineRecon.

\paragraph{Murre \cite{guo2025multiviewreconstructionsfmguidedmonocular}.}
We use the pre-trained checkpoint and default setup to generate depth estimates.
The following TSDF fusion is mostly standard, but we lower the fusion consistency threshold from \(3\) to \(2\), since the default threshold frequently removes too much geometry when only a small number of views overlap.
Sparse-depth conditioning is performed using the full available SfM point cloud, independent of the number of RGB frames used for reconstruction.
For the synthetic input points, we apply view-based visibility masking to mimic the SIFT visibility tracks available for SfM outputs.

\paragraph{MonoSDF \cite{yu2022monosdfexploringmonoculargeometric}.}
We use the official MonoSDF implementation with Omnidata \cite{eftekhar2021omnidata} depth and normal supervision.
Specifically, we use their ScanNet~\cite{dai2017scannet} configuration and optimize for the full \(200\)k iterations.

\paragraph{FineRecon \cite{stier2023finerecondepthawarefeedforwardnetwork}.}
We use the official weights, which were trained on ScanNet.
The released inference pipeline requires scene bounds, which we compute from the ground-truth mesh axis-aligned bounding box, enlarged by \(20\%\).


\newpage

\end{document}